\pdfoutput=1
\documentclass[runningheads]{llncs}
\usepackage{lmodern} 
\usepackage{graphicx, alltt, url}

\usepackage{fancyhdr}
\fancypagestyle{preprint}{
  \fancyhf{} 
  \fancyhead[C]{\footnotesize This work is a preprint for submission early 2026.}
}

\begin{document}
\title{Evaluating Metrics for Safety with LLM-as-Judges}
\titlerunning{Evaluating Metrics for Safety with LLM-as-Judges}

%
\author{Kester Clegg\inst{1} \and 
Richard Hawkins\inst{1} \and 
Ibrahim Habli\inst{1} \and 
Tom Lawton\inst{2}}
\authorrunning{K. Clegg et al.}
%
\institute{Centre for Assuring Autonomy, University of York, York, UK\\
\email{\{kester.clegg, richard.hawkins, ibrahim.habli\}@york.ac.uk} \and
Bradford Royal Infirmary, Bradford, UK\\
\email{Tom.Lawton@bthft.nhs.uk}}
\maketitle              

\thispagestyle{preprint} 

\begin{abstract}
 LLMs (Large Language Models) are increasingly used in text processing pipelines to intelligently respond to a variety of inputs and generation tasks. This raises the possibility of replacing human roles that bottleneck existing information flows, either due to insufficient staff or process complexity. However, LLMs make mistakes and some processing roles are safety critical. For example, triaging post-operative care to patients based on hospital referral letters, or updating site access schedules in nuclear facilities for work crews. If we want to introduce LLMs into critical information flows that were previously performed by humans, how can we make them safe and reliable? Rather than make performative claims about augmented generation frameworks or graph-based techniques, this paper argues that the safety argument should focus on the type of evidence we get from evaluation points in LLM processes, particularly in frameworks that employ LLM-as-Judges (LaJ) evaluators. This paper argues that although we cannot get deterministic evaluations from many natural language processing tasks, by adopting a basket of weighted metrics it may be possible to lower the risk of errors within an evaluation, use context sensitivity to define error severity and design confidence thresholds that trigger human review of critical LaJ judgments when concordance across evaluators is low.
\end{abstract}

\section{Background}
\label{sec:use-of-llms}
\subsection{Assessing the performance of LLMs}
\label{introLLMs}

Productivity gains from using LLMs in domains such as safety analysis are hard to quantify due to the context dependence when assessing performance. Prior work using LLMs for fault tree analysis \cite{clegg-llm-fta} showed that both qualitative and quantitative assessments are challenging due to the reliance on subjective expert opinion and the variable contexts of failure modes within a system. For example, is omitting a base event in a fault logic tree more critical than applying incorrect failure logic at a conjunction? Context dependence can mean that not all errors in the output are of equal severity even when they are of the same type, i.e. \textit{where} the error occurs (its context) may be as critical as the error type. Understanding how to evaluate LLM outputs in a robust, context-sensitive manner lies at the heart of whether we can deploy LLMs to process information flows in critical scenarios. 

If we want to argue that an LLM deployment will not contribute to a known hazard, we need convincing evidence about safeguarding output quality from the model, i.e. can errors be detected and handled, and if not, can they be mitigated? For critical tasks, there may be limited or even no prior data on the types of errors that can be made, or their severity, as critical tasks tend to be performed by suitably qualified or experienced personnel (SQEP) who ensure high quality standards. But when we try to evaluate LLM performance against what good looks like, we need to do more than simply compare LLM outputs to human performance, we need to ensure we can also detect and handle errors that are unique to LLMs. The context sensitivity problem means that detecting these sorts of errors might require greater contextual understanding than the model can be provided with as part of its inputs or prompt for the task, therefore additional safeguards are necessary. The `burden of proof' moves to the robustness of the evaluation points in the process, regardless of whether that evaluation is by machine or human. By safer in this context, we are not talking about an intrinsic quality of the model itself, i.e. how well it aligns to guardrails or is able to resist jailbreaks. We are interested in whether \textit{we can trust an LLM to do a task safely, given the context and requirements of that task}, and what evidence would give us the confidence to argue that. 

Building additional processing `scaffolding' around an LLM is now commonplace to try to meet performance guarantees. Techniques like Retrieval Augmented Generation (RAG), knowledge graphs, fine tuning or over-fitting on local domain data, agentic frameworks, prompt and context engineering have all been claimed to reduce model hallucination and errors. However, there is little agreement on how to evaluate these claims, and due to the context sensitivity problem, generic performance benchmarks carry no weight in most safety critical contexts (see references in \cite{lim2025matrixmultiagentsimulationframework}). To some extent the method chosen to achieve the quality of performance needed is irrelevant, provided that evidence from evaluations demonstrates that the LLM is safe to use for the specific task under consideration in its particular usage context.

However, many of the architectural proposals can be affected by or hide unexpected complexity in their evaluations. For example, sparse exposure to training data (typical of restricted private domains) can mean LLM outputs show higher variance on the same task and small changes in task complexity can have a disproportionately adverse effect on output quality. But `complexity' across different text processing tasks has no recognised metric\footnote{I am excluding human linguistic or cognitive performance measures, such as syntactic depth or semantic / pragmatic loads, as there is no evidence these apply in the same way to LLMs.}, and given that simply increasing input token lengths results in greater misalignment \cite{levy2024task}, how can we assess the likelihood and probable error severity for a given task? As Chen et al. argue, context is essential for judging LLM performance \cite{chen2024premise}, but can we argue that we have sufficient confidence in the evaluation metrics to be able to replace human oversight for a critical task?

\subsubsection{RAG and Knowledge Graphs in Safety-Critical Domains}
\label{prev-work}
In an extensive survey of the capabilities of LLMs to generate safety cases, Odu et al.~\cite{odu2024automaticinstantiationassurancecases} used a combination of linguistic and vector-based metrics to try and assess the LLMs responses' proximity to ground truth. The first was a simple one-to-one token match, the second used BLEU (Bilingual Evaluation Understudy) score \cite{bleu}, which is a metric used to evaluate the quality of machine-translated text by comparing it to human-translated reference text, and thirdly cosine similarity, which is a measure used to compare vector embeddings for words in LLMs. There are weaknesses with all of these metrics, as they do not measure `meaning' as understood by humans with respect to a ground truth. Secondly, such measures do not measure missing information that should have been included in an output, i.e. where an LLM has omitted relevant information (sometimes referred to as the Gricean maxim of quantity). Omission remains an outstanding problem for LLM evaluation, and despite recent work by Microsoft \cite{metropolitansky2025towards} to date the only means to check for omission is to go back to the source material and review it against the LLM output by hand.

Many of the concerns around factual accuracy have been explicitly raised in a recent technical memorandum from NASA looking at the use of LLMs to assess or process safety assurance cases \cite{graydon2025examining}. The report cites a heavily publicised paper from the University of Glasgow that was the first to describe LLMs as ``bullshitters [sic]''  \cite{Townsen-bullshit}, based on the philosopher Harry Frankfurt's definition of someone who has ``no concern for truth or how things really are'' \cite{Frankfurt}. Despite the fact that LLMs are poor at reasoning and ``there is no concept of truth in the LLM training process'' \cite{graydon2025examining}, LLMs often attempt to reason without any understanding of what is true or warning the user that their knowledge of the domain is poor. The NASA report further questions the interactive and repetitive mechanisms employed by many researchers to either improve LLM answers or deal with variance in the LLM output. For example they cite the example of \cite{odu2024automaticinstantiationassurancecases} who asked the same query of an LLM five times and took an average of the answers using a BLEU score in order to deal with output variance. They questioned how this approach could be deemed an efficient or even realistic way to evaluate outputs in production. 

Breaking up large tasks into subtasks is a common tactic to improve outputs from LLMs, and recent developments have seen the providers of frontier models recognise this by developing agentic frameworks \cite{li2024reviewprominentparadigmsllmbased}. Agentic frameworks permit tasks to be allocated to individual agents, and each of those agents can be prompted using system instructions to adopt a specific role, and potentially even fine-tuned to give expertise in a specific context. Model Context Protocol (MCP) developed by Anthropic\footnote{\url{https://docs.anthropic.com/en/docs/agents-and-tools/mcp}} takes this a step further by using a common interface that  standardises how applications provide context to LLMs, allowing LLM agents to start applications to provide input and outputs to the LLM. MCP and similar frameworks often rely on JSON schemas to ensure data is correctly formatted between the LLM and associated applications. This has lead to some vendors providing APIs that guarantee correctly structured JSON output by limiting token outputs from a restricted set\footnote{\url{https://platform.openai.com/docs/guides/structured-outputs}}. While being correctly structured does not mean the content is correct with respect to a ground truth, it could help in certain contexts by ensuring all agents use the same vocabulary when passing information between them. 

Recent work (2021–2025) on retrieval-augmented generation (RAG) and knowledge graphs (KGs) in high-assurance domains highlights domain-specific performance challenges and evaluation difficulties. In healthcare, there are claims that RAG can boost accuracy and reduce hallucinations by grounding LLM outputs in vetted literature or guidelines \cite{Wada2025RadiologyRAG}. For example, a local radiology RAG system eliminated all hallucinated contrast-dosing suggestions present in an LLM’s output (reducing errors from 8\% to 0\%). However, the performance of RAG in such specialized settings hinges on the quality and relevance of the retrieved knowledge base. If the retrieval corpus is outdated or imprecise, RAG may surface irrelevant or low-quality information (“retrieval noise”) that degrades output fidelity \cite{Neha2025RAGHealthcare}. Domain specificity also poses challenges: models can falter when faced with institution-specific terminology or shifting guidelines (a form of domain shift)  \cite{Neha2025RAGHealthcare}. Indeed, integrating newly updated clinical protocols into RAG systems remains non-trivial, and few studies have tested RAG in time-sensitive clinical decision support scenarios \cite{Wada2025RadiologyRAG}. One recent radiology study showed that even with RAG improvements, a local 11B parameter model still lagged behind larger cloud LLMs on overall performance metrics \cite{Wada2025RadiologyRAG}, underlining the difficulty of closing the gap in critical tasks.

Evaluating RAG-driven systems in safety-critical contexts is itself a major challenge. As previously mentioned, standard NLP metrics e.g. BLEU, ROUGE (Recall-Oriented Understudy for Gisting Evaluation), often fail to capture the nuanced requirements of domains like medicine. Research has found that such automatic scores align poorly with human expert judgments \cite{Fukui2025NuclearMedRAG}. For instance, in a nuclear medicine Q\&A system, string-based metrics (ROUGE / Levenshtein distance) did not correlate with radiologists’ assessments of answer quality \cite{Fukui2025NuclearMedRAG}. An LLM-based evaluation metric (RAGAS) achieved more consistent rankings with the experts, but still only modestly correlated to manual scores, implying that expert oversight remains essential. More broadly, RAG evaluation in high-stakes settings often requires task-specific criteria (e.g. factual correctness, clinical relevance, reasoning transparency) and domain-expert panels to judge outputs. The literature points to inconsistent definitions (e.g. what constitutes a “hallucination” and can it be formally defined? \cite{cossio2025comprehensivetaxonomyhallucinationslarge}) and the lack of heterogeneous evaluation protocols across studies \cite{Neha2025RAGHealthcare}, complicating comparisons. 

Researchers have explored knowledge graphs as a tool to enhance reliability and interpretability of AI in safety-critical systems. KGs offer a structured representation of verified facts, which can be used to constrain or inform an LLM’s generation. In complex domains (like aerospace or nuclear operations), this approach aims to enforce logical consistency and provide explainable reasoning paths \cite{Neha2025RAGHealthcare}. In medical decision support, graph-augmented RAG pipelines have shown gains in precision and factual accuracy by linking patient queries to trusted knowledge bases of clinical relationships \cite{Neha2025RAGHealthcare}. By guiding the model’s logic via graph relations rather than unconstrained text, these systems can reduce spurious reasoning and increase interpretability, a critical advantage for high-assurance use. 

Despite these benefits, the use of KGs introduces its own performance challenges. A knowledge graph must be comprehensive, current, and correct --- properties that are hard to guarantee in domains that rely on synchronous real-time data of the type typical in medical settings. In practice, graph-augmented models struggle if the underlying structure is incomplete or out-of-date \cite{Neha2025RAGHealthcare}. Given the scale of proprietary domains, the graph relations are often automatically extracted by LLMs. Maintaining and validating such KGs is labour-intensive, and any mistakes in the graph can propagate to model outputs. There are also other scalability concerns: integrating graph queries with neural generation can be computationally heavy and complex to engineer\cite{Neha2025RAGHealthcare}. These constraints pose questions about the generalizability and real-time use of KG-based methods in critical environments.

Finally, evaluating KG-augmented systems is challenging because success must be measured not just by output accuracy, but by the soundness of the reasoning. This is often done using `chain-of-thought' judgements by LLMs, despite widespread recognition of LLMs' poor reasoning capabilities cited earlier. Traditional metrics may not capture whether an answer was arrived at via valid, trustable knowledge. Domain experts often need to review the provenance of the model’s decisions (e.g. which graph nodes were used) to concretely assess safety. This is reflected in recent medical studies where expert panels conducted multi-factor evaluations (correctness, justification quality, safety implications) for graph-informed LLM outputs \cite{Neha2025RAGHealthcare}.

In summary, the methods discussed bring stringent requirements for knowledge quality, and they expose limitations in our current evaluation methods. Particularly in critical sectors like healthcare (where patient safety is at stake), even minor performance lapses can have serious consequences. There is a clear gap in fully understanding how these systems perform on edge cases and how to guarantee their reliability over time as the augmentation context is updated \cite{Neha2025RAGHealthcare}. While RAG and KGs can mitigate some weaknesses of stand-alone LLMs, the challenge of rigorously evaluating and assuring their performance in safety-critical contexts remains a significant problem.

\subsubsection{LLM-as-Judge in Safety-Critical Contexts}

The problems associated with evaluation bottlenecks needing human review has led to proposals to use LLMs as automated evaluators (LLM-as-Judge or LaJ). Machine evaluation has implications in safety-critical contexts. In addition to what has already been discussed, LaJs may be prone to issues of bias, reproducibility, and variance in their judgments, and there is the issue of `who judges the judge?'. While LLM-based evaluators have been tested as surrogates for expert panels in healthcare and safety assurance, studies consistently show low reliability when domain expertise is required \cite{Fukui2025NuclearMedRAG}\cite{Zhang2024EvaluationBias}, which means some mechanism is required to flag when human review is required to check the model's output. For instance, in nuclear medicine QA tasks, LLM-judged metrics only partially correlated with radiologists’ assessments, raising concerns about entrusting judgment to models that cannot fully grasp context-specific error severity \cite{Fukui2025NuclearMedRAG}. Bias amplification is a recurring issue: evaluators inherit the training data biases of the underlying LLM, leading to uneven scoring across subdomains \cite{Zhang2024EvaluationBias}. The problem is compounded by instability under prompt variations and adversarial phrasing, which undermines reproducibility of evaluations in regulated domains \cite{Ammanabrolu2024EvalSurvey}. Even when LLMs provide rationales for their judgments, these rationales are often post-hoc and cannot be trusted as causal explanations, limiting their use in assurance arguments for critical systems \cite{Liu2023TrustLLM} (see also our later discussion on evaluating metrics for safety in Section \ref{sec-eval-metrics}). 

The use of LaJ extends to classification type automated evaluations. For example, in \cite{lim2025matrixmultiagentsimulationframework} an LLM is used to identify safety relevant dialogues with patients by validating against expert clinician annotations. While this study claims the best LLMs ``out performed clinicians in a blinded assessment of 240 dialogues'', LaJ-based systems such as this require a rigorous evaluation of the LLM as judge to make the safety case sufficient, and these may rely on properties intrinsic to the model. Such properties can be difficult to verify, especially for large models and limited amounts of testing. It is worth mentioning that most of the LaJ use cases that report adequate performance in safety critical contexts tend to be classification type tasks, and as such do not involve risks of omission. Where omission is a substantial risk, the use of RAG or LaJ based augmented evaluation remains insufficient. For example, in \cite{walker2025raguardnovelapproachincontext} they found that even with an extensive augmentation framework (RAGuard and SafetyClamp) around the LLM, the risks of omission in the LLM outputs remained despite being substantially reduced. The assurance argument in these cases rests on whether exposure to the risk of omission is low, or if the exposure is high (i.e. it could occur in many contexts), then the safety case should argue that an omission will not contribute to known hazards. This implies that a method for detecting the omission in LLM output can be assured.

Finally we would draw attention to work on why chain-of-thought reasoning (including LLM-as-a-judge evaluation) fails in medical text understanding \cite{wu2025chainthoughtfailsclinical}. Pertinent to our work, they developed an error taxonomy (hallucination, omission, incompleteness) evaluated via an LLM-as-a-Judge framework with clinician review, offering actionable guidance for safer use of chain-of-thought (CoT) in clinical contexts. They used ROUGE averages and Jacquard index (see also our discussion on this metric for LaJ evaluation in Section \ref{sec-eval-metrics}) to assess tasks like summarisation. What is imporant to note is that they found ``LLM-as-a-Judge framework with expert review also highlights the hallucination and information omission as dominant failure modes of CoT'' and that they ``observe that numbers, units, and measurement-related tokens are overrepresented among CoT-incorrect traces (e.g., pt, ct, numeric values), as are abbreviations for laboratory and physiologic assessments such as plt (platelet count) and pft (pulmonary function test)''. As we show later, these are exactly the types of text items we are trying to assure in our experiment (see Section \ref{sec-implementation}). However, in our case we are not testing model performance. Our purpose is to evaluate what arguments can be put forward for assuring the safety of LaJ use in a critical setting. Whether such arguments are deemed sufficient will depend on what type of evidence is presented to back up those claims and who is making the safety assessment.

Having discussed the literature on methods to ground LLMs in domain data and some of the issues associated with LaJ evaluations, we will now use an illustrative implementation of an agentic framework and examine in detail how LaJ evaluations could provide evidence for assurance. We provide a worked example of the metrics typically used in evaluation points, how they can be composed and discuss what design decisions could be made to assure the output from a safety perspective.

\section{LaJ Use Case: An agentic framework for Peri-operative Risk Assessment }
\label{usecase}

Our objective is to determine what is required from evaluation points to demonstrate to stakeholders (patients, health trusts), users (clinical care professionals) and developers that an LLM-based application is safe to use for a safety-related task. To explore this we use an example of a realistic text processing task within a critical setting. The aim is not to present the assurance of a fully implemented system, but to discuss how a LaJ based agentic system could use evaluation points to argue that the application is safe. To illustrate the problem of context sensitivity described earlier, we chose the medical domain where the error severity is dependent on its context. The patient's condition and treatment can provide multiple overlapping contexts for risk assessment and much of the information that is stored is text based. 

The medical domain is composed of multi-disciplinary teams that come together to work on a patient's surgical or other care needs in what are termed `clinics'. This type of team work, comprising of domain specialists and a co-ordinator is a good fit for an agentic LLM workflow. For these reasons we decided to implement a simple agentic framework, with help from a clinical lead for critical care and anaesthesia, for peri-operative risk assessment clinic that could demonstrate a realistic LaJ task in a safety critical context.

\subsection{Patient safety under anaesthetic}
\label{laj}
\begin{figure}
    \centering
    \includegraphics[width=0.75\linewidth]{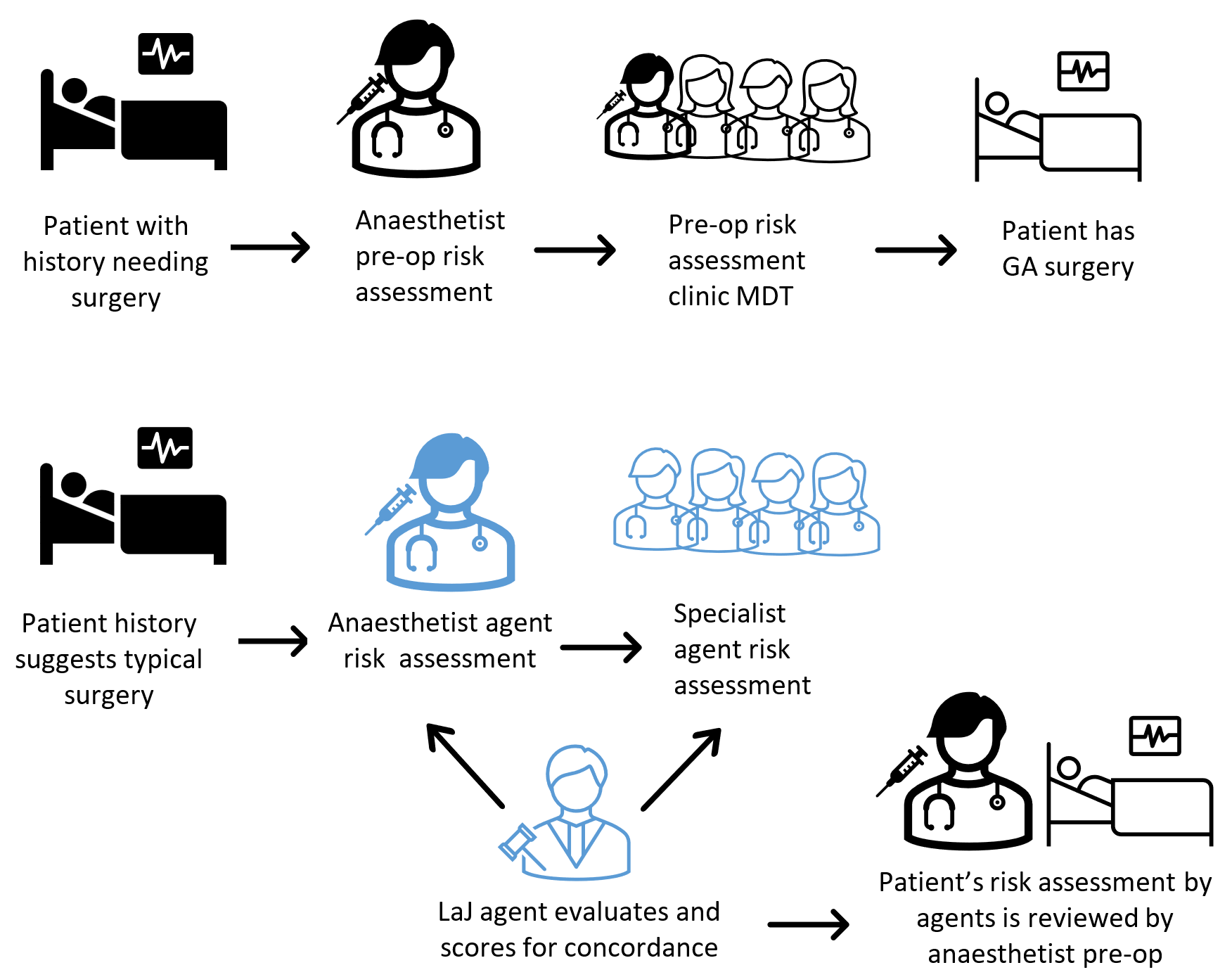}
    \caption{Top diagram shows the workflow for pre-operative risk assessment clinic. Bottom diagram shows our imaginary agentic use case. The surgical context for the patient is provided by the surgeon, or surgeon agent (see Fig. \ref{fig:gpt5anaesthetist}).}
    \label{fig:workflow}
\end{figure}

We first discuss how a patient might be harmed in our chosen context. If a surgeon recommends surgical intervention for a patient, they will call on an anaesthetic pre-assessment clinic to prepare an initial risk assessment based on the patient context, surgery type, estimated blood loss, length of operation, etc. At the clinic, staff will examine the patient record and prepare an initial risk assessment or brief. If needed, they may call on a consultant anaesthetist or other specialist clinician to advise on any unusual or complex conditions that present risks to the patient during surgery. However, it is possible for the clinic to miss some aspect of the patient's condition or underestimate a complex risk scenario. An example of this is a patient who has, at some point in the past, received bleomycin (a chemotherapy agent). Once a patient has been exposed to bleomycin, their lungs may remain sensitised indefinitely and at risk of pulmonary fibrosis if exposed to high levels of oxygen. During general anaesthesia, patients are routinely pre-oxygenated and maintained on inspired oxygen fractions much higher than room air. In patients with a bleomycin history, exposure to high FiO$_2$ (Fraction of Inspired Oxygen) can precipitate or accelerate oxygen-induced pulmonary fibrosis, even years after the chemotherapy was given. Clinically, this may manifest as acute respiratory distress, progressive hypoxaemia, and long-term loss of lung function. Mortality in reported cases is high \cite{ALLAN201252}.

Critical information like this can be difficult to find in the patient's record. Our anaesthetist advisor cites a previous practice of putting a sticker on the front of the patient notes as a visible warning, but in the digital era this often takes the form of one of a multitude of `flags'. While clinical policies vary, the anaesthetist’s risk assessment on the morning of surgery is usually considered the primary check, but pre-operative assessment nurses, surgeons, and oncologists are key safety nets. If bleomycin exposure is missed by the anaesthetist, the most likely prevention would be at the earlier pre-op clinic (by nurses or consultant anaesthetist reviewing the record) or by the oncologist’s treatment summary being reviewed pre-surgery. For oncology patients in particular, MDT (multi-disciplinary team) discussions (oncologist, surgeon, anaesthetist, specialist nurses) may bring prior bleomycin exposure into focus, especially if lung complications are already suspected. 

For our use case, we wanted the LLM to role-play the initial pre-op anaesthetist in preparing a risk assessment brief for a particular patient. For the sake of examining how LaJ agents could improve or even verify LLM outputs, we created an imaginary scenario whereby an \textbf{anaesthetist agent}'s risk assessment is checked against a set of\textbf{ clinical specialist agents}, each of which has been asked to prepare a similar risk brief (see Fig. \ref{fig:workflow} and \ref{fig:gpt5anaesthetist}). Although this does not mirror clinical practice in hospitals, it serves as a realistic context in which we can assess LaJ evaluations in terms of their safety implications. As part of this imaginary scenario, we envisaged a human anaesthetist using the agentic framework to provide an initial risk assessment, knowing it is backed up by LaJ adjudicated comparisons between the anaesthetist agent and specialist agents. The human anaesthetist might assume this combination means that the agentic risk assessment will have no critical errors or omissions. In this case study we consider the argument and evidence that might be put forward to support a safety claim regarding a patient risk assessment generated by LLM agents, given that if the risk assessment were to miss a condition like bleomycin exposure, serious harm to the patient could occur.

\subsection{Implementation}
\label{sec-implementation}
\begin{figure}[ht]
    \centering
    \includegraphics[width=0.9\linewidth]{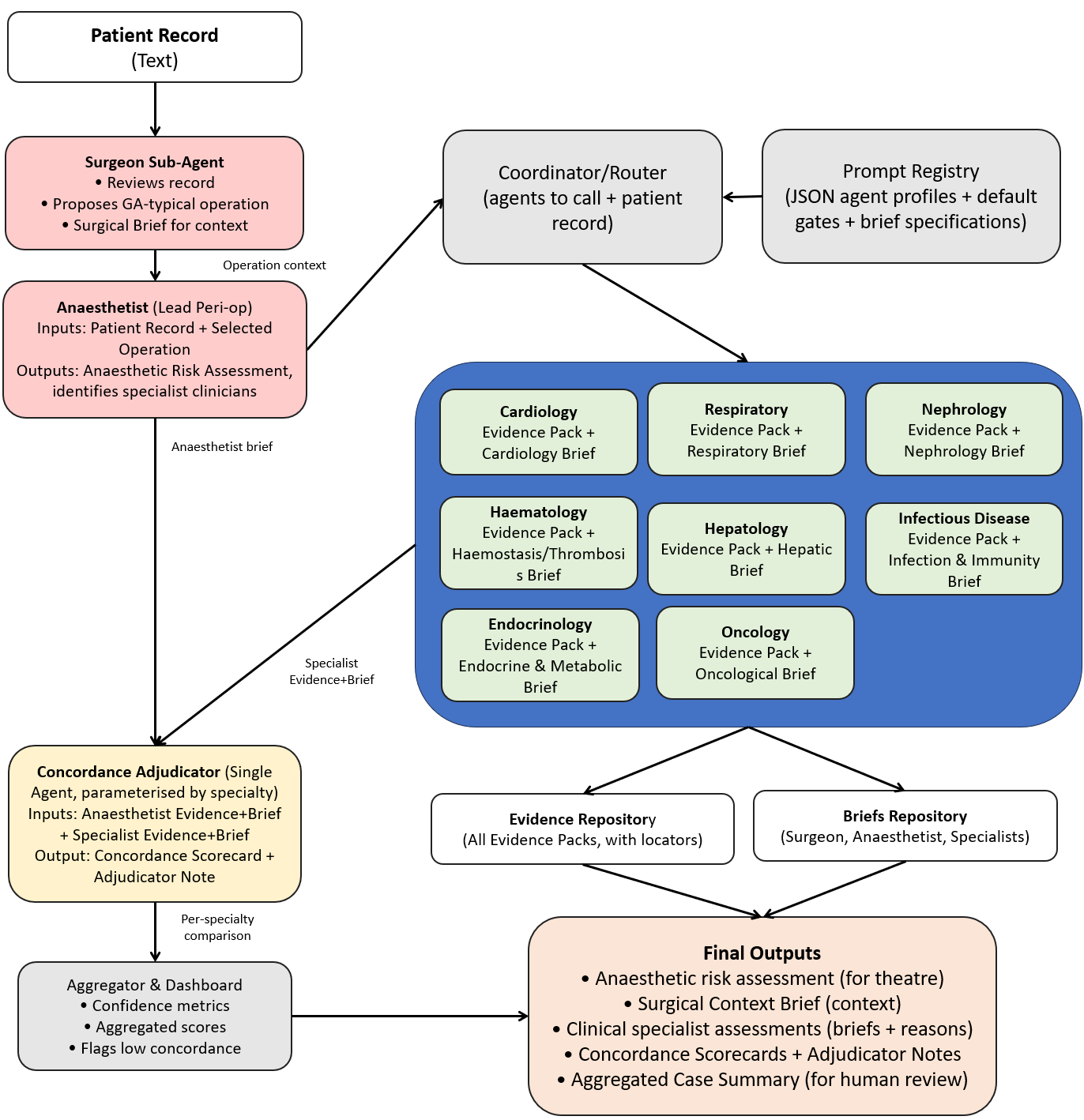}
    \caption{The workflow for peri-operative risk assessment clinic performed by LLM agents.}
    \label{fig:gpt5anaesthetist}
\end{figure}
The agentic framework we created uses individual system prompts based on profiles for clinical roles. Each agent has this stored as JSON file, in which there are explicit instructions relating to quality gate assessments, risk brief profiles and evidence briefs (verbatim extracts from the patient records). 


Our main purpose in implementing this agentic framework was to look at a real-world example of context sensitivity for error types and detection, and how evaluations by LaJs require careful composition in order to lower the risks of unsafe judgments by the LaJ adjudicator. The LaJ evaluation described below is not intended to generate debate on whether these are the most suitable metrics for this task. Rather, we are interested in how these types of evaluation points can be assured, or help assure the safety of the final outputs. We do not present statistical results of how often errors were detected in an agent's output or how often the anaesthetist's brief was found to have missed critical contexts, as these are model dependent. Instead we discuss how to make evaluation metrics evidence-based and traceable, and whether non-deterministic or `soft' metrics based on LLM judgements can be trustworthy.

We generated risk categories for the peri-operative risk assessment framework by collecting a small set of patient records and extracting clinical risk factors.\footnote{Patient record set from huggingface: \url{https://huggingface.co/datasets/jung1230/patient_info_and_summary/viewer/default/train?views\%5B\%5D=train&row=35}.} Each patient was individually assessed for risk assuming a common surgical procedure for that patient profile that would require a general anaesthetic. These risks were grouped into a minimal set of common peri-operative risk categories (e.g. cardiovascular / haemodynamic, airway / respiratory, renal, infectious / immunological). This taxonomy provided a consistent way to capture and compare risks across different patients and specialist contexts, ensuring that the categories were consistent and risks could be linked to categorised domains rather than left as free text extractions.

Having ranked and classified the risks, we could identify the clinical specialists that an anaesthetists might call upon for specialist advice or input in a peri-operative meeting. Our framework implements the simple agentic workflow shown in Fig. \ref{fig:gpt5anaesthetist}. The \textbf{Surgeon} agent suggests a plausible surgical intervention appropriate to the patient profile; the \textbf{Anaesthetist} agent then reviews the patient record and proposed operation, producing a structured risk brief and identifying which clinical specialist agents should be consulted. Each \textbf{specialist} agent (e.g. \textbf{Cardiologist}, \textbf{Respiratory}, \textbf{Nephrologist}, \textbf{Haematology}, etc) has a defined profile and prompt tailored to its domain, ensuring that its output is a focused summary of the most critical risks for that speciality together with supporting evidence extracted verbatim from the patient's record.

The \textbf{Adjudicator} agent (i.e. LaJ) then compares the  Anaesthetist agent’s risk assessment with that from each of the specialist agents. The Adjudicator's role is to \textit{evaluate concordance by checking coverage of risks, correctness, prioritisation, and actionability}, with special attention to predefined quality gates (critical domain-specific safety rules). Each gate is checked against the evidence and the Anaesthetist agent's brief; missed gates cap the relevant sub score (potentially triggering human review). Sub scores across the risk categories are weighted and aggregated to produce an overall confidence score (0–100) together with a grade band (High, Medium, Low, Very Low). The adjudicator outputs both a machine-readable scorecard (JSON: sub scores, agreements, disagreements, gate checks) and a short explanatory note, in theory giving a human and machine auditable measure of how well the Anaesthetist agent's risk assessment aligns with specialist agents' expertise.

\subsubsection{Agent profiles}
\label{sec-eval}
Each of the agent's system prompts is defined by a JSON file so that they can be altered and examined outside the program code and structured outputs are easier to generate. Agent profiles can be extensive in real world applications. We reproduce a brief extract from the Cardiology agent system prompt below, as the full profile for this clinical specialist is much longer. It should be noted that in the implementation we had to shorten most of the specialist profiles to prevent agents misaligning during their tasks. This seems to be a case of ``context poisoning'', i.e. where a lengthy contextual input starts to confuse an agent. However, we believe this limitation could be overcome with greater development effort. The evidence packs and risk briefs are defined separately.

\subsubsection*{EVIDENCE PACK Line Structure example (JSONL; one object per extracted item)}
\begin{itemize}
  \item \textbf{source\_id}: string (e.g.\ file name or document id)
  \item \textbf{locator}: string (e.g.\ page/section/line or timestamp) --- be specific
  \item \textbf{extract\_text}: string (VERBATIM; no edits)
  \item \textbf{tag}: one of {\small \texttt{["CAD/ACS","HF", "VALVE", "ARRHYTHMIA", "PULM\_HTN", "AORTA", "ANTITHROMBOSIS", "FUNCTION", "INVESTIGATION", "LABS", "COMORBID"]}}
  \item \textbf{comment}: string, $\leq 12$ words 
\end{itemize}
\noindent Example line:
\begin{small}
    \begin{verbatim}
    {"source_id":"preop_note.pdf",
     "locator":"p3 §2 lines 12–18",
     "extract_text":"Severe aortic stenosis… AVA 0.7 cm², 
        mean gradient 45 mmHg.",
     "tag":"VALVE",
     "comment":"critical AS metrics"}
    \end{verbatim}
\end{small}

\subsubsection*{Extract from Cardiology Risk Brief}
\begin{enumerate}
  \item \textbf{Top Risks (CVS Context)}  
  5--8 bullets, highest-impact first.
  \item \textbf{Immediate Actions / Optimisation}  
  Concrete steps (e.g., arterial line on induction, hold clopidogrel 5 days if safe, consider bridging per haematology).
  \item \textbf{Go / Delay Triggers}  
  Explicit criteria that should delay surgery (e.g., decompensated HF, ongoing troponin rise, critical AS without plan).
  \item \textbf{Monitoring / Adjuncts}  
  e.g., intra-operative TEE, invasive monitoring, device magnet policy, post-operative ICU/HDU plan.
  \item \textbf{Risk Scoring (if data available)}  
  e.g., EuroSCORE II / STS (cardiac); procedure-specific vascular risk; otherwise state ``UNKNOWN''.
\end{enumerate}

\subsubsection*{Quality Gates (before finalisation)}
\begin{itemize}
  \item Re-check numbers digit-by-digit and units.
  \item Confirm that every brief bullet is supported by at least one Evidence Pack line.
  \item No clinical instruction conflicts (e.g., holding antiplatelet in recent stent without plan) --- flag conflicts in ``Go/Delay Triggers''.
\end{itemize}
\subsubsection*{Additional Quality Gates (Cardiology)}
\begin{itemize}
  \item \textbf{ACS / ongoing ischaemia}: if symptoms \emph{or} rising troponin $\Rightarrow$ list as first Top Risk;  
  for non-emergent vascular surgery include a delay trigger; for cardiac surgery, specify revascularisation/operative strategy as per evidence.
  \item \textbf{Stents / DAPT}: Evidence Pack \textbf{must} include stent type + date + indication + all antiplatelets with last doses.  
  Brief \textbf{must} state peri-operative antiplatelet plan or MDT decision if within high-risk window (BMS $<$30 d; DES $<$3--6 mo)\\
  \ldots \textit{(remainder cut short)}
\end{itemize}
The definition of these agent profiles could be worked out with the peri-operative team. In particular, the quality gates could be defined to suit local patient profiles. We shortened the list of critical quality gates in our implementation to allow the LaJ evaluations to be more easily inspected. For example for Nephrology, this was reduced to: gates=[K+ $\geq 6.0$ or ECG changes → delay/correct first", "Platelets <100×10$^9$/L or INR >1.5 → correct before procedure"]. Keeping the list of safety critical quality gates to minimal lengths greatly helped the consistency of the specialist agents' risk briefs.

The adjudicator agent evaluates agreement between the anaesthetist's brief and each specialist brief using a structured rubric. Inputs include the ANAESTHETIST\_BRIEF, SPECIALIST\_BRIEF, supporting verbatim extracts (ANAESTHETIST \_EVIDENCE\_JSONL, SPECIALIST\_EVIDENCE\_JSONL), and specialist QUALITY\_GATES.
\subsubsection{Dimensions of Comparison}
The adjudicator computes sub scores (0--1) across five dimensions (weighting in brackets):
\begin{enumerate}
  \item \textbf{Coverage (30\%)} --- proportion of specialist risk items also present in the anaesthetist brief. This is a Jacquard Index overlap calculation. \emph{Example:} 8 of 10 items matched $\Rightarrow$ coverage = 0.8.

  \item \textbf{Critical Items (30\%)} --- hard rules defined by \texttt{QUALITY\_GATES}.  
  If a gate is supported by evidence but absent in the anaesthetist brief, the subscore is capped:
  \begin{itemize}
    \item One major miss: $\leq 0.40$
    \item Multiple major misses: $\leq 0.20$
  \end{itemize}
  \emph{Example:} LVEF 35\% (gate requires invasive monitoring); if omitted, critical\_items $\leq 0.4$.

  \item \textbf{Correctness \& Specificity (20\%)} --- penalises contradictions or vague statements.  
  \emph{Example:} Evidence: SpO$_2$ 88\% RA; anaesthetist states ``respiratory function stable'' $\Rightarrow$ reduced specificity.

  \item \textbf{Prioritisation Alignment (10\%)} --- compares ordering of top risks.  
  \emph{Example:} Specialist places hypoxaemia as \#1, anaesthetist places it last $\Rightarrow$ subscore reduced.

  \item \textbf{Actionability Alignment (10\%)} --- compares monitoring, optimisation, and delay triggers.  
  \emph{Example:} Specialist: ``Delay if K$^+ \geq 6.0$'' vs anaesthetist: ``Monitor potassium'' (no threshold) $\Rightarrow$ reduced score.
\end{enumerate}

\subsubsection{Disagreements and Quality Gates}
Disagreements in our implementation tended to be mostly misses (in clinical practice, actual disagreements would be rare), but are categorised as:
\begin{itemize}
  \item \textbf{MISS} --- absent in anaesthetist brief but present in specialist evidence.
  \item \textbf{OVERCALL} --- asserted without evidence.
  \item \textbf{CONFLICT} --- direct contradiction.
  \item \textbf{AMBIGUOUS} --- insufficient evidence, should be marked ``unknown''.
\end{itemize}
Each disagreement is recorded with a severity (minor / moderate / major). Critical gates act as hard caps: if a major gate is missed (e.g.\ ``Acute LRTI $\Rightarrow$ delay elective''), the overall score is capped at $\leq 69$ and the case is flagged for human review. In theory this prevents superficial agreement from obscuring dangerous omissions (however, see discussion in Section \ref{sec-eval-metrics}). The key disagreements are shown as coloured badges in the dashboard summary (shown on the RHS of Fig. \ref{fig:dashboard}).

\subsubsection{Grading}
\begin{figure}[ht]
    \centering
    \includegraphics[width=0.8\linewidth]{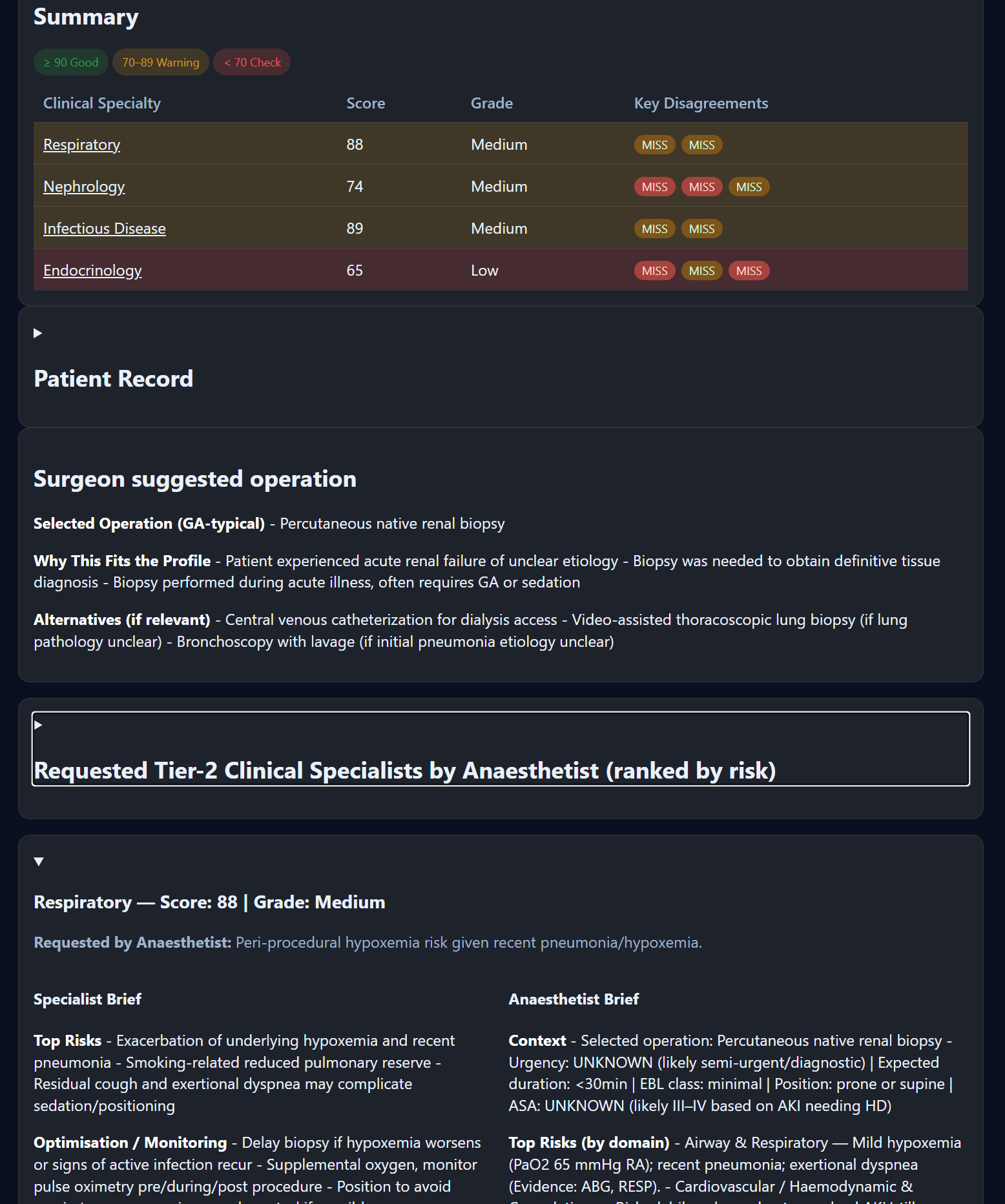}
    \caption{Adjudicator results are summarised via a simple dashboard for Peri-Operative Risk Assessment Clinic.}
    \label{fig:dashboard}
\end{figure}

The overall score is a \textbf{weighted sum} that is mapped to a concordance grade as follows: High: $\geq 90$;  Medium: 70--89; Low: 40--69; Very Low: $<40$. Fig. \ref{fig:dashboard} shows a dashboard display of the concordance scores, in a summary table with one row per specialist, coloured according to the concordance grade (from green (high agreement), to amber, to red (low or capped)). 

\subsubsection{Dashboard summary}
The web-based dashboard (Fig. \ref{fig:dashboard}) summarises our imaginary peri-operative risk assessment clinic for a patient undergoing surgery under general anaesthetic. The interface displays the patient record, the surgeon agent’s suggested operation, and the anaesthetist agent’s requested specialist reviews (ranked in order of clinical risk). Comparisons are shown in drop down cards, with each card containing the specialist agent’s brief, the anaesthetist agent’s brief, structured evidence extracts, and an adjudicator-generated scorecard for each of the dimensions listed above. The summary table at the top provides traffic-light colour coding of agreement scores and clickable links to specific disagreements. Traceability of assessment judgments can be accessed through verbatim evidence links and structured sub scores, therefore every claim in the briefs maps back to verbatim extracted JSONL lines from the patient record (source id + locator + extract text). However, in practice the verbatim extracts were not found easy to match with the briefs due to the complex text sometimes present in a patient's record, meaning that human reviews could take time to confirm the assessment was correct.

\section{Evaluating metrics for safety}
\label{sec-eval-metrics}
We have described how the risk categories and agent profiles were developed, and have shown examples of how the concordance scores made by the Adjudicator LaJ claim to be calculated. However, we have not considered the Adjudicator's evaluations from the perspective of safety. If this were a real-world implementation, it would be classified as a medical device and have to undergo a full safety audit. The question we want to pose is whether the concordance scores could lead to harm for the patient, i.e. could a high concordance score still miss a critical factor in the patient record, like exposure to bleomycin, leading to patient harm under anaesthetic care?  

When assuring LaJ based applications, it is essential to know what the limits of the LLMs are with regard to evidence, and what is claimed as part of the evaluation. For example, if we take a metric like \textbf{coverage} of critical risk factors using the quality gates comparison, and decide the Jacquard Index is an adequate measure, we need to ensure a) that the LLM implements this metric correctly, and b) where could errors in that metric occur and could they be detected? There is a risk if the LLM is asked to create the metric, it will propose it but not declare it explicitly in the prompt. Thus in the explanation of how the \textbf{coverage} figure is determined, the LLM claims: \textit{coverage = (specialist items matched) / (total specialist items)}. But when we examine the LLM derived prompt, it does not make this explicit: \textit{COVERAGE (0.30): \% of Speciality's salient risks present in BRIEF\_S that BRIEF\_A also captures (allow synonymous phrasing)}. Of course, we can explicitly prompt it to follow the maths, but will it actually execute that? Does it pass it on to a deterministic calculator (external tool or generated code) rather than relying on next token prediction? 

But even if we insist on the maths to use and force it to deploy a calculator or executable program code, are we sure that the specialist risk items are being correctly identified and matched? Medical abbreviations or chemical symbols are common in specialist domains but their usage varies \cite{wu2025chainthoughtfailsclinical}. For example, instead of quoting verbatim from the patient record, the anaesthetist's brief rewrote ``arterial oxygen saturation'' as $SpO_2$. This is one of several possible variants for this synonym in the patient records. What evidence can be provided that all variants will be identified as synonyms? Perhaps this last issue brings us back to deploying a common list of quality gates agreed by clinical specialists so that agents deploy from a restrictive JSON schema under the STRICT output flag. This would guarantee a level of conformance to one vocabulary by the agents, but not necessarily over the input (i.e. the patient record). A list of possible synonyms might give greater assurance over input variance, but you have to demonstrate that the LLM will use that list by testing against the full range of possible inputs. The evidence-based arguments for coverage apply equally to the identification of \textbf{critical items}, including how errors might be generated by the LLM or the adjudicator agent when calculating the score.

Softer metrics, like \textbf{correctness \& specificity} need additional care if used in evaluations. Although an LLM may claim it is performing something that looks deterministic, closer examination proves this has little clarity. When asked to provide examples of \textbf{correctness} it gave the following: \textit{If anaesthetist says “stable CAD, no DES” but specialist evidence shows a DES → penalise}. But the prompt states: \textit{CORRECTNESS \& SPECIFICITY (0.20): Are statements in BRIEF\_A factually supported by EVIDENCE\_PACK\_A (and not contradicted by EVIDENCE\_PACK\_S)? Penalise overcalls and inaccuracies.} There is no definition of what ``penalise'' means in terms of quantity, the range of contexts it could apply and whether it is proportional to the contextual risk. Likewise, no evidence can be obtained about what ``factually supported'' means in terms of the metric. The LaJ is generating a numerical score based on vaguely defined judgements that directly affect the concordance grade, which is intended to be a safety assured metric.

For \textbf{specificity} the model gave the example: \textit{If anaesthetist says “low oxygen” without numbers, while evidence gives $SpO_2$ = 88\%→ penalise specificity}. While we do not claim that specificity \textit{per se} is a poor metric to use, we would raise the question about whether it is a good metric in every context. The anaesthetist's assessment of low oxygen in this context is correct and knows what that means for the surgical procedure. Increasing the specificity does not lower risk for the patient in this context. Instead it might raise false positives for poor concordance causing unnecessary human review. In terms of safety however, this metric is presented as part of basket of measures that assure the quality of the risk assessment comparison. But perfect specificity only means for those risks identified, there was close agreement. Risks that are not identified would not have the metric applied (i.e. an undetected error), giving a false impression about the quality of the risk assessment comparison. The safeguard of capping a concordance score if a critical risk has been missed by the anaesthetist agent only partially mitigates this, if both miss the error would be undetected.

When composing a basket of measures, including soft metrics, it is important from a safety perspective to understand whether the weight of one metric could lessen the weight of a more important metric that directly relates to a harmful outcome for the patient. Weighted metrics need to have accompanying evidence that explains the rationale behind the weighting, and an argument verifying that high or low scores in one metric (even if correct) will not undermine the overall safety of the LaJ judgement.

\section{Conclusions}
\label{sec-conclusions}
We have presented an example of a composable, weighted metric of the type typically used to argue that a LaJ-based evaluation can be trusted. The metric seems well designed for a safety critical context: it is risk-based, context sensitive and can be adjusted to align with local clinical policies or patient intake. 

However, parts of the scoring rubric remain shrouded in mystery. We can ask an LLM how it calculates coverage, but that is insufficient evidence to use in a safety claim. An LLM can produce a chain-of-thought explanation of how it arrives at a score, but that does not guarantee what it is doing ‘under the hood’. We can only get that type of evidence if the calculation is passed to an external tool or program code whose calculations we can trace. Unfortunately there are aspects of text based comparisons, for example gaining assurance that synonyms will be recognised equally, or the use of unquantified criteria like ``factually supported'' that are dependent on LLM judgments and which will remain difficult to assure. The value of soft metrics such as these should be questioned, even if it can be shown they are reliably consistent in deployment, so that their contribution to the concordance score is well understood in terms of overall risk. While assigning weighted scores is good practice in LaJ judgements, the reasoning behind the weighting should align with where the greatest risk is for that patient.

Despite attempting to mitigate risks by using evaluation points on the LLM outputs, risks of hallucination, omission and incompleteness by the LaJ itself remain, particularly in medical settings \cite{wu2025chainthoughtfailsclinical}. LaJ is touted as the only way to scale agentic system throughput, as human-in-the-loop agentic workflows quickly bottleneck around the human reviews. However, as we have argued, LaJ based assurance of LLM outputs need to have greater determinism from their metrics if they are to be used in safety critical environments. They need explicit instruction on the quantification of the metric, traceable evidence of what they use and proof of deterministic calculations before we can argue that their judgements can be assured.

\bibliographystyle{abbrv}
\bibliography{cfaa}

\end{document}